%% file: main.tex
\documentclass{article} 
\usepackage{iclr2027_conference,times}

\input{math_commands.tex}

\usepackage{hyperref}
\usepackage{url}
\usepackage{graphicx} 
\graphicspath{{./includes/}, {./includes/supplementary_includes/}}
\usepackage[]{units} 
\usepackage{xspace} 
\usepackage{amsfonts} 
\usepackage[dvipsnames]{xcolor} 
\usepackage{subcaption}  
\usepackage{float}
\usepackage{amsmath} 
\usepackage{bbm}     
\usepackage[T1]{fontenc}
\usepackage{algorithm}
\usepackage{algpseudocode}
\usepackage{caption}
\usepackage{booktabs}
\usepackage{multirow}
\usepackage{tabularx}
\usepackage{amssymb}
\newcolumntype{L}{>{\raggedright\arraybackslash}X}

\title{Game-Guided Skill Discovery through \\ Self-Play for Playable Agent Control}

\author{Seungeun Rho$^{1}$ \qquad
Jeonghwan Kim$^{1}$ \qquad
Xue Bin Peng$^{2}$ \qquad
Sehoon Ha$^{1}$ \\[0.5em]
$^{1}$Georgia Institute of Technology \\
$^{2}$Simon Fraser University and NVIDIA \\[0.3em]
\texttt{\{srho31,jkim3662,sehoonha\}@gatech.edu},
\texttt{xbpeng@sfu.ca}
}

\usepackage[commandnameprefix=ifneeded]{changes}
\definechangesauthor[name={Sehoon Ha}, color=blue]{SH}
\setdeletedmarkup{}              
\renewcommand{\deleted}[2][]{}   

\iclrfinalcopy 
\begin{document}

\maketitle

\begin{abstract}
We present \textbf{Game-Guided Skill Discovery (GGSD)}, a framework that uses self-play in games to discover motor skills that are directly playable by humans. Playable skills provide a compact abstraction for controlling embodied agents through a small set of learned behaviors rather than low-level actions. To be effective, these skills should be semantically distinct, interpretable, and expressive; properties that existing unsupervised skill-discovery methods often fail to achieve simultaneously. GGSD achieves these desiderata by grounding skill discovery in competitive gameplay. A hierarchical agent competes against its past selves, with a high-level policy selecting from a small discrete skill set and a skill-conditioned low-level policy learning the corresponding behaviors. After training, a human can replace the high-level policy and directly control the agent through the same discrete skills. Despite the small number of high-level actions, skill transitions give rise to emergent \emph{combo behaviors}, expanding expressivity beyond individual primitives. Across Ant, Franka-arm, and Unitree G1 environments, we show that GGSD produces human-playable skills that humans can compose to solve unseen tasks, such as \texttt{Maze} and \texttt{CubePush}, without additional training. An interactive demo is available at \url{https://ggsd-demo.github.io}.

\end{abstract}

\begin{figure*}[h!]  
    \centering
    \includegraphics[width=0.85\textwidth]{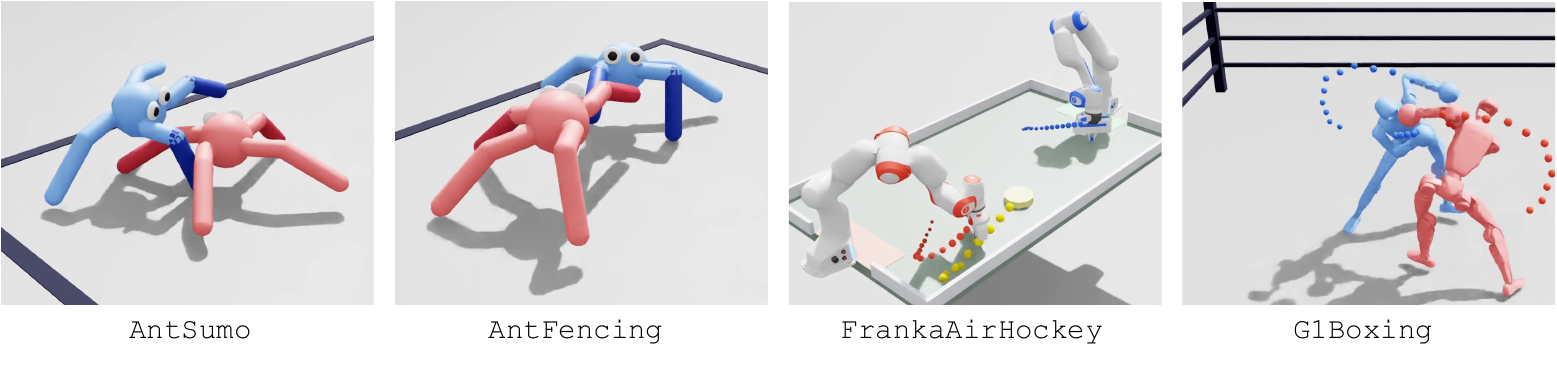}
    \caption{
Different game rules guide agents toward different playable skills.
Shown are gameplay snapshots of agents learned with \methodabbr{}.
}
    \label{fig:teaser}
\end{figure*}

\section{Introduction}

Learning a compact repertoire of reusable motor skills is a long-standing goal in skill discovery. Such skills can provide useful abstractions not only for downstream policy learning, but also for direct human control: if complex behaviors are organized into a small set of intuitive skills, a human can select and compose them without specifying low-level actions. For this interface to be effective, the discovered skills should be semantically distinct, interpretable, scalable to high-degree-of-freedom agents, and sufficiently expressive despite a compact action space.

Existing unsupervised skill-discovery methods do not naturally guarantee these properties. Most approaches optimize objectives based on mutual information (MI) \citep{gregor2016variational,sharma2019dynamics,eysenbach2019diversity, kwon2020variational, laskin2022unsupervised} or Wasserstein dependency measures (WDM) \citep{lsd,park2024metra}, encouraging different skills to induce distinguishable state distributions or trajectories. While this can produce interpretable behaviors in simple environments, increasing agent complexity introduces many ways for skills to differ without being behaviorally meaningful. As a result, distinctiveness alone can yield skills that are easy to distinguish but difficult to interpret or reuse.

We argue that \emph{competitive games} provide a natural source of structure for discovering such skills, as illustrated across diverse embodiments in Figure~\ref{fig:teaser}. A simple game rule specifies what constitutes success while leaving open how success should be achieved. Through self-play, evolving opponents continually expose the agent to new strategic and physical situations, encouraging the emergence of useful behaviors without requiring users to specify individual skills a priori. Self-play has long been shown to induce emergent behaviors, from superhuman strategies in competitive games \cite{silver2016mastering,silver2018general,vinyals2019grandmaster,baker2020emergent,oh2021creating} to structured motor behaviors in physically embodied agents \cite{bansal2018emergent,haarnoja2024learning,jansonnie2024unsupervised}. We ask whether this same mechanism can be used to discover a compact repertoire of skills that is directly playable by humans.

Based on this idea, we introduce \textbf{Game-Guided Skill Discovery (\methodabbr)}, a framework that uses self-play in simple 1v1 competitive games to discover playable motor skills. Each agent is controlled by a hierarchical policy. A high-level policy selects from a small set of discrete skills, while a skill-conditioned low-level policy maps the selected skill to motor actions. The game objective encourages behaviors that are useful for winning, while a mutual-information objective encourages different skill codes to acquire distinct behavioral semantics.

After training, we remove the high-level policy and map each skill to a human input, such as a keyboard button, allowing users to directly control the agent without training a new task-specific controller. We intentionally keep the number of skills small (only five or six) in all environments to maintain a simple interface for human play. Despite this compact action vocabulary, the controller gains additional expressivity through \emph{skill transitions}: executing one skill can place the agent in a state from which another skill produces a qualitatively different behavior, giving rise to emergent \emph{combo behaviors}. Similar to button combinations in commercial games, these transitions allow a small set of discrete skills to support behaviors richer than the individual primitives alone.

As a result, \methodabbr{} discovers semantically diverse and human-interpretable skills even for high-degree-of-freedom embodiments such as humanoids. These skills are also directly reusable beyond the games in which they are learned. Without any additional policy training, humans can compose them to solve previously unseen downstream tasks, including complex locomotion in \texttt{Maze} and object interaction in \texttt{CubePush}.

The core contributions of our work are as follows:
\begin{itemize}
\item We introduce \methodabbr{}, a skill-discovery framework that uses self-play in games as lightweight guidance for learning human-playable motor skills.

\item We show that \methodabbr{} discovers semantically distinct and human-interpretable skills that scale to high-degree-of-freedom agents. Despite using only a small discrete skill set, transitions between skills give rise to emergent \emph{combo behaviors} that substantially expand the controller's expressivity.

\item We demonstrate \methodabbr{} across Ant, Franka Arm, and Unitree G1 environments, and show that humans can directly compose the learned skills to solve previously unseen locomotion and object-interaction tasks without additional policy training.

\end{itemize}

\section{Related Works}

\subsection{Self-Play for Motor Skill Learning}

Fictitious play provides a classical mechanism for stabilizing self-play by repeatedly learning best responses to the empirical average of opponents' historical strategies~\citep{brown1951iterative,robinson1951iterative}. Fictitious self-play extends this idea to extensive-form games \citep{heinrich2015fictitious}, while Neural Fictitious Self-Play (NFSP) approximates best responses with deep reinforcement learning and historical average strategies with supervised learning \citep{heinrich2016deep}. These methods primarily aim to learn equilibrium strategies. Similarly, our method trains against a distribution of historical policies. Competing against a diverse pool of past selves encourages the agent to develop reusable skills that remain useful across a wide range of opponents rather than specializing to a particular strategy.

Consistent with this intuition, self-play and competitive interaction have been shown to induce complex motor behaviors in physically simulated agents. \citet{bansal2018emergent} demonstrated the emergence of behaviors such as running, blocking, tackling, and kicking through multi-agent competition, while later works extended competitive learning to bipedal soccer \citep{haarnoja2024learning}, robotic manipulation \citep{jansonnie2024unsupervised}, and hierarchical multi-drone volleyball \citep{zhang2025mastering}. Won et al.~\citep{won2021control} studied high-DoF humanoids in boxing and fencing, but learned the underlying motor skills from reference motions before training competitive strategies.

In contrast, we use self-play itself as guidance for \emph{discovering} the low-level skill repertoire. Rather than learning a monolithic game-playing policy or relying on predefined/reference-based motor skills, our method explicitly organizes behaviors induced by competition into discrete, diverse, and reusable skills that can be directly controlled by humans.

\subsection{Guidance in Skill Discovery}

Several works introduce external guidance to address a key limitation of unsupervised skill discovery, as optimizing solely for distinctiveness or state coverage can produce behaviors that are diverse but semantically meaningless. Language-Guided Skill Discovery~\citep{rho2025language} uses LLM-generated state descriptions to encourage semantically distinct skills. However, obtaining language descriptions throughout the explored state space becomes increasingly impractical as agent dimensionality grows. DoDont~\citep{kim2024s} uses desirable and undesirable demonstrations, while Reference Grounded Skill Discovery~\citep{rho2026reference} uses reference motions to guide the learned skill repertoire. While effective for high-DoF agents, the discovered repertoire is largely shaped by the provided motion dataset, and preparing sufficiently diverse reference motions can itself be costly. In contrast, guidance from simple game rules is lightweight. A game's rules can be specified in only a few lines of code, while self-play autonomously discovers how to succeed under those rules. This allows the agent to discover novel behaviors beyond explicitly provided examples, while remaining scalable to high-DoF systems.

\section{Preliminaries}

\paragraph{Markov Decision Process.}
We first consider a standard Markov decision process (MDP), defined by
$\mathcal{M}=(\mathcal{S},\mathcal{A},P,r,\gamma)$, where
$\mathcal{S}$ and $\mathcal{A}$ denote the state and action spaces,
$P(s' \mid s,a)$ is the transition probability,
$r(s,a)$ is the reward function, and
$\gamma \in [0,1)$ is the discount factor.
A policy $\pi(a\mid s)$ is trained to maximize the expected discounted return
\begin{equation}
    J(\pi)
    =
    \mathbb{E}_{\pi,P}
    \left[
        \sum_{t=0}^{\infty} \gamma^t r(s_t,a_t)
    \right].
\end{equation}
\paragraph{Two-player games.}
\label{sec:preliminaries}
We consider a two-player game between an agent and an opponent.
The game state is given by $s = (s^{\mathrm{me}}, s^{\mathrm{foe}}),$
where $s^{\mathrm{me}}$ denotes the agent's own state, and
$s^{\mathrm{foe}}$ denotes the state of the opponent.
Similarly, the two players take actions
$a^{\mathrm{me}} \in \mathcal{A}^{\mathrm{me}}$
and
$a^{\mathrm{foe}} \in \mathcal{A}^{\mathrm{foe}}$.
The game dynamics are described by
\begin{equation}
    P_{\mathrm{game}}
    \left(
        s'
        \mid
        s,
        a^{\mathrm{me}},
        a^{\mathrm{foe}}
    \right),
\end{equation}
and the controlled agent receives a game reward
$r^{\mathrm{me}}(s,a^{\mathrm{me}},a^{\mathrm{foe}})$.

When the opponent follows a fixed policy
$\pi_{\mathrm{foe}}(a^{\mathrm{foe}}\mid s)$ over continuous actions,
its action can be marginalized into the environment dynamics.
In particular, we define the induced transition probability
\begin{equation}
    \tilde{P}_{\pi_{\mathrm{foe}}}
    (s' \mid s,a^{\mathrm{me}})
    =
    \int_{\mathcal{A}^{\mathrm{foe}}}
    \pi_{\mathrm{foe}}
    (a^{\mathrm{foe}}\mid s)
    P_{\mathrm{game}}
    \left(
        s'
        \mid
        s,
        a^{\mathrm{me}},
        a^{\mathrm{foe}}
    \right)
    \, da^{\mathrm{foe}},
\end{equation}
with the corresponding expected reward
\begin{equation}
    \tilde{r}_{\pi_{\mathrm{foe}}}
    (s,a^{\mathrm{me}})
    =
    \int_{\mathcal{A}^{\mathrm{foe}}}
    \pi_{\mathrm{foe}}
    (a^{\mathrm{foe}}\mid s)
    r^{\mathrm{me}}
    (s,a^{\mathrm{me}},a^{\mathrm{foe}})
    \, da^{\mathrm{foe}}.
\end{equation}
Therefore, for a fixed opponent policy, the two-player game reduces to an
ordinary single-agent MDP
\begin{equation}
    \mathcal{M}_{\pi_{\mathrm{foe}}}
    =
    \left(
        \mathcal{S},
        \mathcal{A}^{\mathrm{me}},
        \tilde{P}_{\pi_{\mathrm{foe}}},
        \tilde{r}_{\pi_{\mathrm{foe}}},
        \gamma
    \right).
\end{equation}
From the perspective of the controlled agent, training against a fixed
opponent is thus equivalent to standard reinforcement learning in an
environment whose dynamics implicitly include the opponent's behavior.
Self-play can be viewed as repeatedly changing this MDP as the
opponent policy evolves over the course of training.

\section{Game-Guided Skill Discovery}

\begin{figure*}[t]  
    \centering
    \includegraphics[width=0.95\textwidth]{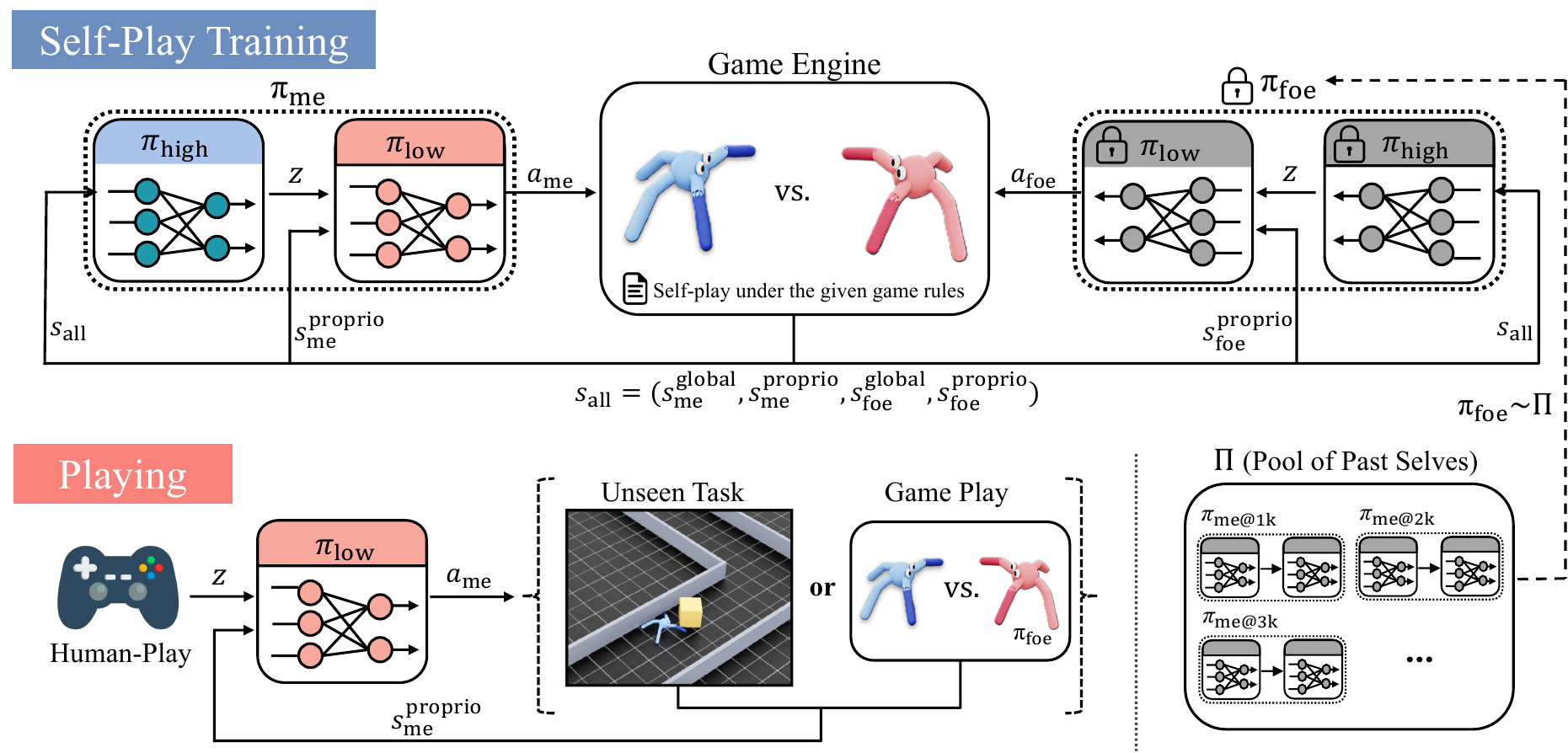}
    \caption{
Overview of \methodabbr. \textbf{Top:} We jointly learn a hierarchical policy through self-play against opponents sampled from a pool of past checkpoints. \textbf{Bottom:} After training, a human can replace the high-level policy and directly control the agent across diverse scenarios.}
    \label{fig:method_overview}
\end{figure*}

\methodabbr{} converts game rules into human-playable skills through self-play. We first describe our self-play procedure, then introduce hierarchical skill learning, and finally explain how the learned skills are made directly playable by humans.

\subsection{Self-Play with a Pool of Past Selves}
\label{sec:self_play}

A straightforward form of self-play trains against a copy of the current policy.
However, as the policy is updated, the opponent changes simultaneously, resulting
in a continuously moving training objective.
Instead, as illustrated in Figure~\ref{fig:method_overview}, we maintain a pool of
historical policy checkpoints and sample a fixed opponent at the beginning of each
episode.

Recall from Sec.~\ref{sec:preliminaries} that a fixed opponent policy
$\pi_{\mathrm{foe}}$ induces a single-agent MDP
$\mathcal{M}_{\pi_{\mathrm{foe}}}$.
We maintain an opponent pool
\begin{equation}
    \Pi_K = \{\pi^{(1)},\ldots,\pi^{(K)}\},
\end{equation}
where $\pi^{(K)}$ denotes the most recent checkpoint.
When the pool size $K$ is greater than 1, we sample an opponent according to
\begin{equation}
    \rho_K(k)=
    \begin{cases}
        p, & k=K, \\[3pt]
        \dfrac{1-p}{K-1}, & 1 \leq k < K,
    \end{cases}
\end{equation}
and keep it fixed throughout the episode.
Thus, the latest checkpoint is sampled with probability $p$, while the remaining
probability is distributed uniformly across earlier checkpoints.

This results in an opponent distribution that is piecewise stationary rather than continuously
changing, while still maintaining competitive pressure from recent opponents.
The parameter $p$ controls the balance between these two effects: larger values
place greater emphasis on the latest opponent, while smaller values increase
diversity from past strategies.
We use $p\in[0.6,0.8]$ in all experiments.

Our procedure is inspired by fictitious play
\citep{brown1951iterative,robinson1951iterative}
and its self-play variants
\citep{heinrich2015fictitious,heinrich2016deep}.
While we do not implement exact fictitious play, we follow its central principle
of training against a distribution of past strategies, with additional emphasis
on recent opponents.

\subsection{Hierarchical Skill Learning}

\paragraph{Jointly learning strategies and motor skills.}
\methodabbr{} jointly learns the high- and low-level policies during self-play. The high-level policy determines \emph{which} skill to execute
as part of the current game strategy, while the low-level policy determines
\emph{how} that skill is physically realized. Joint training allows the two
levels to co-evolve: improved strategies expose the agent to new situations that call for new low-level skills, while an expanding skill repertoire enables more sophisticated strategies.

This joint training is implemented through a temporal hierarchy. Let
$s_t^{\mathrm{all}}=(s_t^{\mathrm{me}},s_t^{\mathrm{foe}})$ denote the full
game state, and let $\tau_j=jk$ denote the $j$-th high-level decision time.

The high-level policy selects a discrete skill
\begin{equation}
    z_j \sim
    \pi_{\mathrm{high}}
    \left(
        z \mid s_{\tau_j}^{\mathrm{all}}
    \right),
\end{equation}
which is held fixed for the following $k$ environment steps. Within this
interval, the low-level policy produces a motor action 
at every step,
\begin{equation}
    a_t \sim
    \pi_{\mathrm{low}}
    \left(
        a \mid s_t^{\mathrm{proprio}}, z_j
    \right),
    \qquad
    \tau_j \leq t < \tau_j+k,
\end{equation}
where $s_t^{\mathrm{proprio}}$ contains only the controlled agent's proprioceptive observations.

\paragraph{Hierarchical policy optimization.}
During training, we retain the sampled skill variables and optimize the joint
distribution over the augmented trajectory. For one skill interval, its
policy-dependent probability factorizes as
\begin{equation}
\begin{aligned}
&p_\theta
\left(
    z_j,
    a_{\tau_j:\tau_j+k-1}
    \mid
    s_{\tau_j:\tau_j+k-1}
\right)
=
\pi_{\mathrm{high}}
\left(
    z_j \mid s_{\tau_j}^{\mathrm{all}}
\right)
\prod_{t=\tau_j}^{\tau_j+k-1}
\pi_{\mathrm{low}}
\left(
    a_t \mid s_t^{\mathrm{proprio}},z_j
\right).
\end{aligned}
\label{eq:hierarchical_factorization}
\end{equation}
Thus, the joint log likelihood decomposes into high- and low-level terms,
allowing the two policies to be optimized at their respective temporal
resolutions. We use separate PPO objectives for both levels, following prior
work on joint hierarchical policy optimization~\citep{li2020subpolicy}.

Each level maintains its own value function and reward. The high-level policy is optimized only for the game objective, since it is
responsible for strategic skill selection.
The low-level policy additionally receives the mutual-information reward $r_t^{\mathrm{MI}}$ defined in Eq.~\ref{eq:mi_reward}, which encourages different skill codes
to acquire distinct motor behaviors:
\begin{equation}
    r_t^{\mathrm{high}}=r_t^{\mathrm{game}},
    \qquad
    r_t^{\mathrm{low}}
    =
    r_t^{\mathrm{game}}
    +
    \lambda_{\mathrm{MI}}r_t^{\mathrm{MI}}.
\end{equation}

We view the game objective as providing the behavioral guidance that shapes which skills emerge, while MI primarily associates these behaviors with distinct skill codes; see Appendix~\ref{app:mi_perspective} for further discussion.

The low-level critic
$V_{\mathrm{low}}(s_t^{\mathrm{all}},z_j)$ operates at every environment
step. The high-level critic operates only at skill-selection steps, and is trained using target values calculated according to:
\begin{equation}
    \label{eq:high_level_target}
    y_j^{\mathrm{high}}
    =
    \sum_{i=0}^{k-1}
    \gamma^i r_{\tau_j+i}^{\mathrm{game}}
    +
    \gamma^k
    V_{\mathrm{high}}
    \left(
        s_{\tau_{j+1}}^{\mathrm{all}}
    \right).
\end{equation}
Thus, the high-level policy uses an effective discount factor of $\gamma^k$ between consecutive skill decisions, while the low-level policy uses the original discount factor $\gamma$ at every environment step.
We use $k=10$ in all experiments.

\subsection{Playable Learned Skills}

Having described how the two levels are jointly trained, we now introduce the
design choices that allow the learned low-level skills directly playable by
humans.

\paragraph{Discrete and diverse skills.}
We use a discrete skill variable $z\in{1,\ldots,M}$, represented as a one-hot vector, with $M=5$ or $6$ depending on the environment. This compact skill set provides a simple interface for human control, while state-dependent skill transitions enable emergent \emph{combo behaviors} (Sec.~\ref{sec:qualitative_analysis}).

To prevent the high-level policy from collapsing to a small subset of skills, we regularize it with categorical entropy:
\begin{equation}
    \mathcal{H}_{\mathrm{high}}
    =
    \mathbb{E}_{s^{\mathrm{all}}}
    \left[
        \mathcal{H}
        \left(
            \pi_{\mathrm{high}}(\cdot \mid s^{\mathrm{all}})
        \right)
    \right].
\end{equation}
This encourages the policy to leverage the full discrete skill set during gameplay.

\paragraph{Mutual-information maximization.}
High-level entropy encourages diverse skill \emph{selection}, but does not
ensure that different codes correspond to distinct behaviors. We therefore
maximize the mutual information between the selected skill $Z$ and the
resulting state $S$. A discriminator $q_\phi(z\mid s)$ gives the variational
lower bound
\begin{equation}
    I(Z;S)
    \geq
    \mathcal{H}(Z)
    +
    \mathbb{E}_{p(z,s)}
    \left[
        \log q_\phi(z\mid s)
    \right].
\end{equation}
Although the skill distribution is induced by the high-level policy, it is
fixed with respect to $\theta_{\mathrm{low}}$ during the low-level update.
The discriminator therefore provides the intrinsic reward
\begin{equation}
    \label{eq:mi_reward}
    r_t^{\mathrm{MI}}
    =
    \log q_\phi(z_j\mid s_t).
\end{equation}
Together, high-level entropy encourages the use of multiple skills, while the
MI objective encourages different skill codes to induce distinguishable
behaviors.

\paragraph{Opponent-agnostic low-level control.}
For direct human control, each skill should retain consistent motor semantics
across different game situations. To encourage this, we restrict both the
low-level policy and the discriminator to proprioceptive observations, while
allowing the high-level policy to observe the full game state:
\begin{equation}
    \pi_{\mathrm{high}}
    \left(
        z\mid s^{\mathrm{all}}
    \right),
    \qquad
    \pi_{\mathrm{low}}
    \left(
        a\mid s^{\mathrm{proprio}},z
    \right),
    \qquad
    q_\phi
    \left(
        z\mid s^{\mathrm{proprio}}
    \right).
\end{equation}
This places opponent-dependent strategic reasoning in the high-level controller
and prevents the discriminator from distinguishing skills based on opponent
states, encouraging low-level skills to retain consistent motor semantics that
are easier for humans to control.

Putting these components together, the two policies optimize the following objective:
\begin{equation}
\begin{aligned}
\theta_{\mathrm{high}}^*
&=
\argmin_{\theta_{\mathrm{high}}}
\mathcal{L}_{\mathrm{high}}^{\mathrm{PPO}}
\left(r^{\mathrm{game}}\right)
-
\lambda_{\mathrm{H}}\mathcal{H}_{\mathrm{high}}, \\
\theta_{\mathrm{low}}^*
&=
\argmin_{\theta_{\mathrm{low}}}
\mathcal{L}_{\mathrm{low}}^{\mathrm{PPO}}
\left(
r^{\mathrm{game}}
+
\lambda_{\mathrm{MI}}r^{\mathrm{MI}}
\right).
\end{aligned}
\label{eq:objective}
\end{equation}
Pseudocode for \methodabbr{} is provided in Algorithm~\ref{algorithm:ggsd}, and the full hyperparameters in Appendix~\ref{app:hyperparam}.

\section{Experiments}

\subsection{Implementation Details}

\paragraph{Game Rules.}
\methodabbr{} transforms a given game rule into a set of playable motor skills.
We consider four games with distinct objectives and embodiments.
(1) In \texttt{AntSumo}, an agent wins by pushing its opponent out of the arena, while falling results in a loss.
(2) In \texttt{AntFencing}, an agent wins by touching the opponent's root body with one of its front legs; falling or leaving the arena results in a loss.
(3) In \texttt{FrankaAirHockey}, two Franka robot arms compete to strike a puck into the opponent's goal.
(4) In \texttt{G1Boxing}, two humanoid robots compete by striking the opponent's head or torso, with the objective of inflicting more damage than the opponent.
Detailed reward terms, observations, and environment configurations for each game are provided in the Appendix~\ref{app:env_details}.

\paragraph{Self-Play.}
Sampling a separate opponent policy for each of the 4,096 environments is prohibitively expensive in GPU memory, so we use grouped opponent sampling.
We divide the 4,096 parallel environments into 10 groups, each of which samples one opponent policy shared by all environments in the group.
Thus, training requires only the current policy and 10 frozen opponent policies to reside on the GPU simultaneously.
We add the current policy to the opponent pool every 2,000 updates and resample each group's opponent every 200 updates.

\subsection{GGSD Learns to Play Games Well}

\begin{figure*}[t]
\centering

\begin{minipage}[t]{0.65\textwidth}
    \vspace{0pt}
    \centering
     \includegraphics[width=\linewidth]{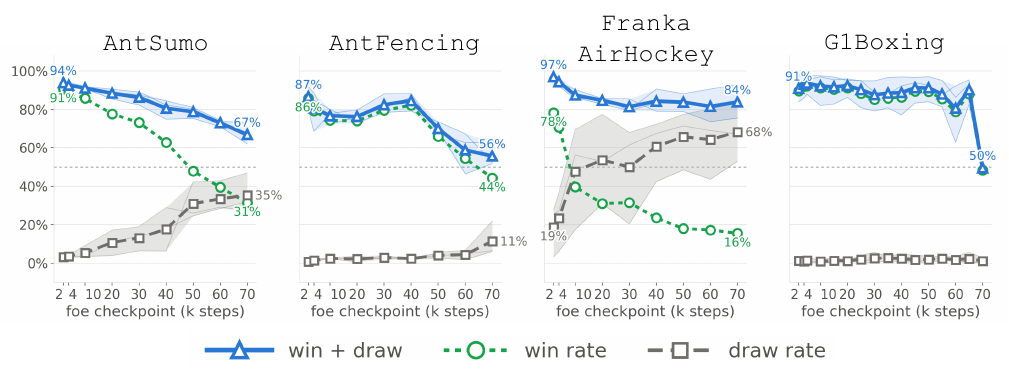}
    \captionof{figure}{Win and draw rates of the final policy against itself and its previous checkpoints.}
    \label{fig:win_rate_vs_pastselves}
\end{minipage}
\hfill
\begin{minipage}[t]{0.32\textwidth}
    \vspace{0pt}
    \centering
    \includegraphics[width=\linewidth]{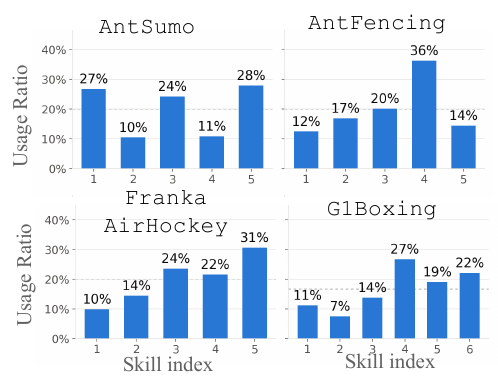}
    \vspace{-1.65em} 
    \captionof{figure}{Skill usage ratios, showing no collapse.}
    \label{fig:skill_usage_ratio}
\end{minipage}
\vspace{-1em}  
\end{figure*}

We first examine whether our self-play procedure is stable and whether it produces agents that can effectively play the given games.
Figure~\ref{fig:win_rate_vs_pastselves} reports the performance of the final policy, trained for 70k updates, against checkpoints from 2k to 70k updates. For each of three independent training runs, we evaluate each checkpoint pair over 1,000 games and average the results.

\paragraph{Self-play produces increasingly competitive agents.} The final policy achieves high win rates against policies from early stages of training.
As the opponent checkpoint becomes more recent, the win rate decreases while the draw rate increases, indicating that the policies become increasingly competitive.
Importantly, the combined win-and-draw rate remains above $50\%$ against every evaluated checkpoint.
For \texttt{G1Boxing}, the final policy achieves over $80\%$ win rate against every past checkpoint except itself at 70k, suggesting that training has not yet fully saturated. As training approaches a plateau, stronger recent opponents should lead to a more gradual decline in win rate. 

The learned agent also performs strongly against human players.
Four users each played five games against the final 70k checkpoint on \texttt{AntSumo}, achieving only two human wins ($10\%$ human win rate).
Overall, these results suggest that training improves the policy against the opponent pool as a whole, rather than overfitting to a particular opponent checkpoint.

Figure~\ref{fig:skill_usage_ratio} further shows how frequently each learned skill is selected during gameplay.
Across all games, the high-level policy does not collapse to a single skill; instead, it consistently makes use of multiple skills.
This indicates that the discovered skill set remains behaviorally relevant during competitive play.

\subsection{Qualitative Analysis of Learned Skills}
\label{sec:qualitative_analysis}

\begin{figure*}[t]
\centering
\includegraphics[width=0.9\textwidth]{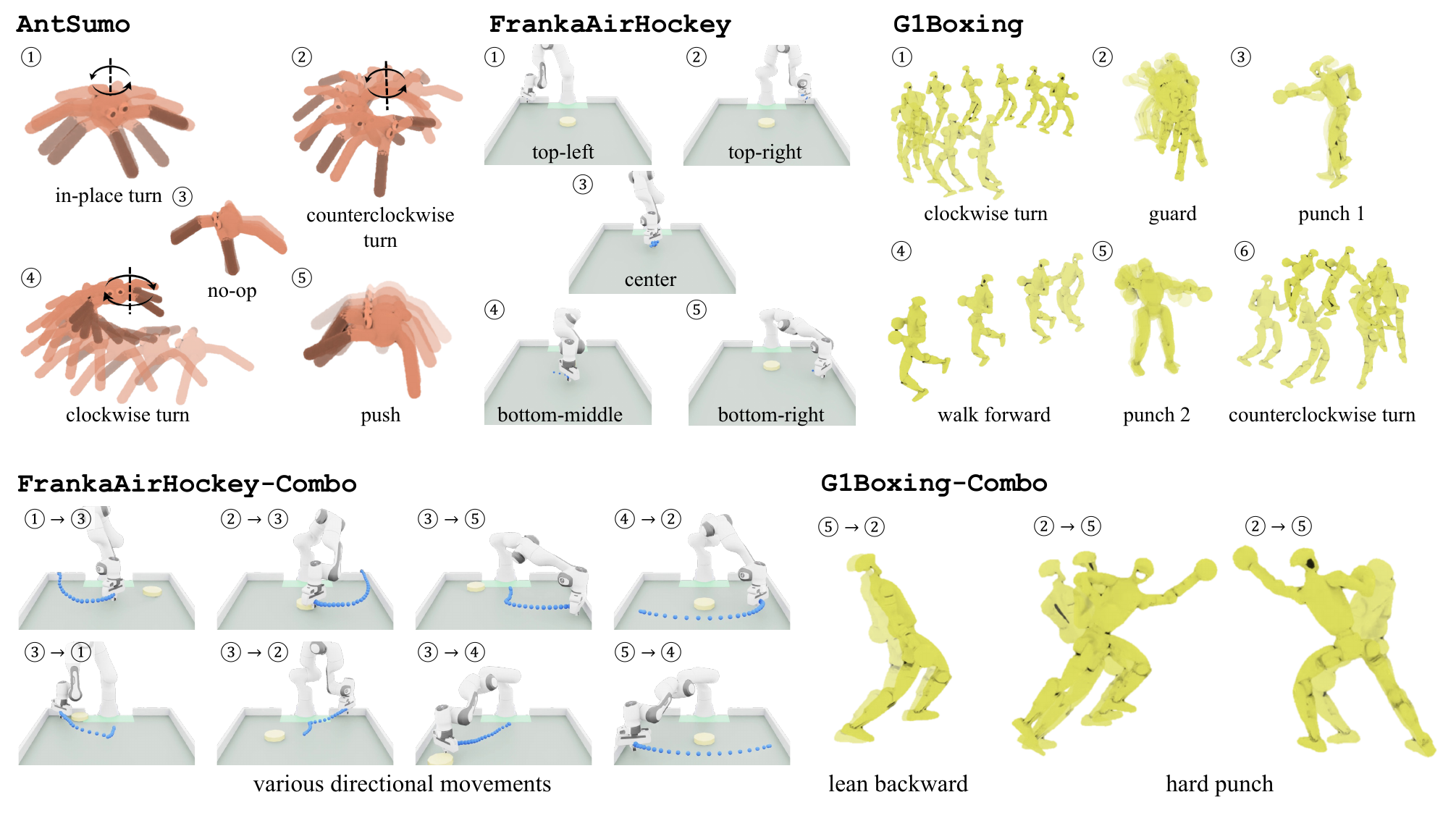}
\vspace{-0.3em}  
\caption{Learned motor skills (top) and combo behaviors through skill transitions (bottom).}
\label{fig:learned_skills}
\end{figure*}

We next examine what motor behaviors emerge from \methodabbr{}.
Figure~\ref{fig:learned_skills} visualizes all skills learned in each game.
For visualization, we fix a single skill at the beginning of an episode, execute it continuously for five seconds, and overlay snapshots of the resulting motion.

\paragraph{GGSD discovers highly interpretable skills.}
The learned skills exhibit clear and readily interpretable behavioral semantics.
For \texttt{AntSumo}, the agents discover a variety of turning, locomotion, and pushing behaviors.
In \texttt{G1Boxing}, the humanoid learns basic locomotion primitives as well as task-specific behaviors such as punching and guarding.
For example, one skill raises the arms in front of the upper body, resembling a defensive guard against incoming punches.

These qualitative patterns are consistent across random seeds.
Although different seeds do not necessarily produce identical skill sets, they repeatedly recover behaviors that are important for successful gameplay, such as locomotion and striking.
This suggests that the game objective provides a consistent behavioral structure while still allowing multiple solutions to emerge.

\paragraph{Skill transitions produce emergent combo behaviors.}
A particularly interesting phenomenon is that meaningful behaviors can emerge not only from individual skills but also from transitions between them.
This is especially apparent in \texttt{FrankaAirHockey}.
Individual Franka skills are relatively static: when executed in isolation, each skill moves the end effector toward a particular region of the table and then remains there.
However, transitions between these discrete skills generate rapid and purposeful movements that are used to strike the puck.

A similar effect appears in \texttt{G1Boxing}.
For example, transitioning from skill 5 to skill 2 produces a backward lean used to evade or absorb punches, whereas the reverse transition produces a strong forward punch by rapidly shifting momentum.
These examples show that discrete skills act as compositional building blocks whose transitions produce richer behaviors than the individual skills alone.

\paragraph{State-dependent skill semantics.}
Emergent combos reveal an interesting tension between semantic consistency and compositional expressivity: the effect of a skill can depend on the state induced by preceding skills. Nevertheless, the resulting controllers remain playable, suggesting that consistency need not hold globally. Instead, each skill can retain predictable semantics within a small number of state regimes, while transitions between regimes enable qualitatively different behaviors. This expands the expressivity of a compact skill set without increasing the number of human inputs, although too many regimes could make the interface difficult to interpret. We discuss this further in Appendix~\ref{app:combo_semantics}.

\subsection{Quantitative Analysis Against Baselines}

\begin{figure*}[t]
\centering
\includegraphics[width=0.9\textwidth]{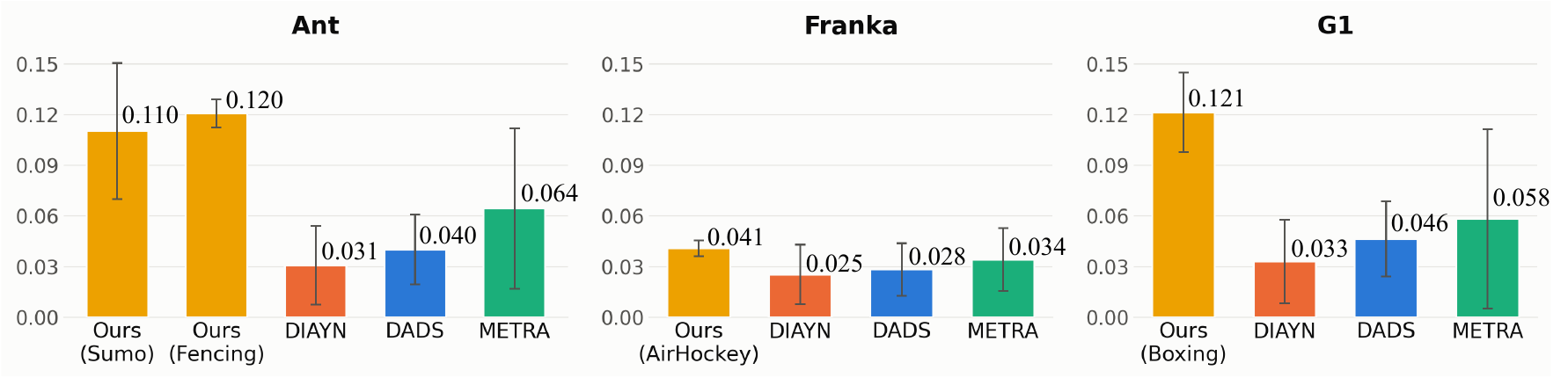}
\caption{
Semantic diversity of learned skills across different embodiments. Higher is better.
}
\vspace{-0.4em}
\label{fig:semantic_diversity}
\end{figure*}

We quantitatively evaluate how effectively game-based guidance promotes \emph{semantically distinct} skills.
Distinctiveness alone is insufficient. Two skills may be distinct while both lack clear semantic meaning.

We therefore adapt the \emph{language distance} metric of \citet{rho2025language} to a vision-language setting.
For each skill, we sample rollout videos at 5~fps, use \texttt{Gemini-3.6-Flash} to generate a natural-language behavior description, and compute pairwise distances between skill descriptions using \texttt{Gemini-Embedding-2}.
For each training run, we average these pairwise distances to obtain a semantic-diversity score, and report the mean across three runs.
Results are averaged over three seeds, with prompting and evaluation details provided in the Appendix~\ref{app:vlm_diversity}.
We compare against DIAYN, DADS~\citep{sharma2019dynamics}, and METRA~\citep{park2024metra}. Since METRA uses a continuous skill space, we train it with a 5-dimensional skill and evaluate five skills corresponding to the one-hot basis vectors.

\paragraph{\methodabbr{} learns semantically distinct skills.}
As shown in Figure~\ref{fig:semantic_diversity}, \methodabbr{} achieves the highest semantic diversity across all embodiments.
It is approximately $2\times$ higher than the strongest baseline on Ant and G1, and 28\% higher on Franka.
These results indicate that game-based guidance promotes not only diverse motions, but semantically distinct behaviors.

\subsection{Human Play on Unseen Downstream Tasks}

\begin{figure*}[t]
    \centering

    \begin{minipage}[t]{0.68\textwidth}
        \centering
        \includegraphics[width=\linewidth]{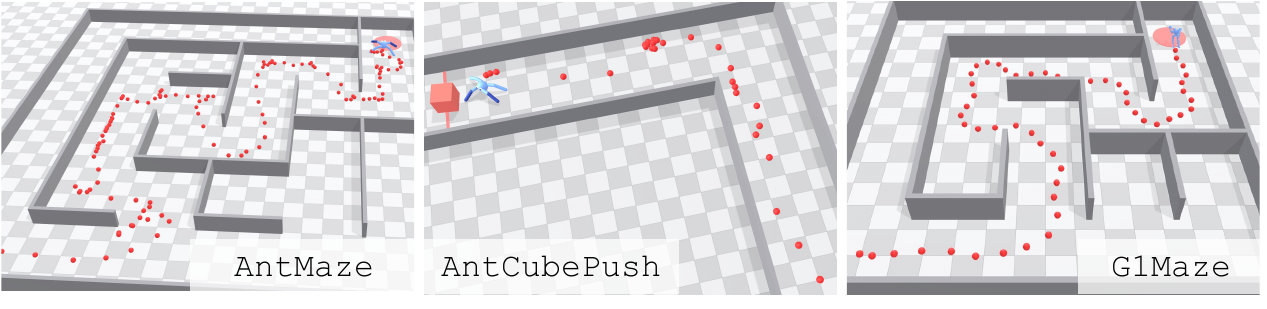}
        \vspace{-1.5em}
        \captionof{figure}{Successful human-play trajectories on unseen tasks. Red dots indicate cube in \texttt{AntCubePush}, and robot in \texttt{Maze}.}
        \label{fig:human_play}
    \end{minipage}
    \hfill
    \begin{minipage}[t]{0.29\textwidth}
    \centering
    
    \vspace{-6.0em}
    \captionof{table}{Human play results.}
    \label{tab:human_play_results}
    \setlength{\tabcolsep}{2pt}
    \begin{tabular}{lcc}
        \toprule
        Task & Succ. & Time(s) \\
        \midrule
        AntMaze & 100\% & 82.7 \\
        AntCube & 100\% & 62.2 \\
        G1Maze & 84\% & 22.0 \\
        \bottomrule
    \end{tabular}
\end{minipage}
\vspace{-1em}  
\end{figure*}

Finally, we evaluate whether skills learned through gameplay can be directly reused by humans on unseen tasks. We keep the learned low-level policies fixed and replace the high-level policy with human input, without additional training. 

Using the policy learned from \texttt{AntSumo}, we evaluate \texttt{AntMaze} and \texttt{AntCubePush}; we also evaluate the \texttt{G1Boxing} policy on \texttt{G1Maze}. In all tasks, users observe a top-down view and select among the learned skills in real time.

\paragraph{GGSD skills are directly playable by humans.}
We conducted a human evaluation with five participants, including two authors, each performing five trials per task.\footnote{This human evaluation was reviewed and approved by an Institutional Review Board (IRB).} Before evaluation, participants read brief control tips and practiced for at most five minutes. Table~\ref{tab:human_play_results} reports the success rate and average completion time across all participants and trials. The average success rate is at least $84\%$ on all three tasks. Figure~\ref{fig:human_play} shows representative trajectories from the user study. \texttt{AntCubePush} requires object manipulation and the maze tasks introduce wall contacts, neither of which is present in the training games. Moreover, all policies are trained in Isaac Lab~\citep{mittal2025isaaclab} but evaluated in MuJoCo~\citep{todorov2012mujoco} web. Despite these task and simulator shifts, humans can compose a small set of GGSD skills to reliably navigate and manipulate objects without additional policy optimization.

\section{Conclusion}

We investigated whether self-play in 1v1 competitive games can produce reusable motor skills that are directly playable by humans. The learned skills support interactive gameplay and can be reused by humans without additional training to solve unseen downstream tasks. Our results suggest that games provide a simple form of guidance for discovering diverse, reusable, and human-playable skills.

\paragraph{Limitations and future work.}
A central limitation of \methodabbr{} is that the learned skill repertoire is bounded by the skills required by the game. Simple games may demand only a narrow set of behaviors, limiting transfer to downstream tasks. For example, a policy trained on \texttt{G1Boxing} is unlikely to acquire dexterous manipulation skills. Scaling \methodabbr{} to games that require broader and more diverse motor capabilities may enable more generic skill repertoires that transfer across a wider range of downstream tasks, which we view as an important direction for future work.


\bibliography{iclr2027_conference}
\bibliographystyle{iclr2027_conference}

\appendix

\newpage

\begin{algorithm}[h!]
\caption{Game-Guided Skill Discovery (\methodabbr)}
\label{algorithm:ggsd}
\begin{algorithmic}[1]
\Require $\pi_{\mathrm{high}}$, $\pi_{\mathrm{low}}$, discriminator $q_\phi$,
opponent pool $\Pi_K$, latest-opponent probability $p$,
skill duration $k$, $\lambda_{\mathrm{MI}}$, $\lambda_{\mathrm{H}}$

\For{each training iteration}
    \State Construct the opponent distribution $\rho_K$ as in
    Sec.~\ref{sec:self_play}
    \State Initialize rollout buffers
    $\mathcal{D}_{\mathrm{high}}$ and $\mathcal{D}_{\mathrm{low}}$

    \For{each rollout environment or environment group}
        \State Sample opponent index $i\sim\rho_K$ and fix
        $\pi_{\mathrm{foe}}\gets\pi^{(i)}$

        \For{each high-level decision step $t$}
            \State Sample
            $z\sim\pi_{\mathrm{high}}
            (\cdot\mid s_t^{\mathrm{all}})$

            \State Execute $z$ for up to $k$ environment steps using
            $\pi_{\mathrm{low}}
            (\cdot\mid s^{\mathrm{proprio}},z)$

            \State Compute per-step MI and low-level rewards
            according to Eqs.~\ref{eq:mi_reward} and~\ref{eq:objective}

            \State Store low-level transitions in
            $\mathcal{D}_{\mathrm{low}}$

            \State Accumulate the discounted game reward over the
            executed skill interval and store the resulting
            high-level transition in $\mathcal{D}_{\mathrm{high}}$
        \EndFor
    \EndFor

    \State Update $q_\phi$ on $\mathcal{D}_{\mathrm{low}}$
    by maximizing
    $\mathbb{E}[\log q_\phi(z\mid s^{\mathrm{proprio}})]$

    \State Update $\pi_{\mathrm{low}}$ and $V_{\mathrm{low}}$
    with PPO using the low-level reward in Eq.~\ref{eq:objective}

    \State Update $\pi_{\mathrm{high}}$ and $V_{\mathrm{high}}$
    with PPO using the high-level target in
    Eq.~\ref{eq:high_level_target}
    and entropy regularization

    \If{checkpoint interval is reached}
        \State Add the current hierarchical policy to the opponent pool
        and increment $K$
    \EndIf
\EndFor
\end{algorithmic}
\end{algorithm}

\section{State-Dependent Skill Semantics and Emergent Combos}
\label{app:combo_semantics}

Our experiments reveal a tension between \emph{semantic consistency} and \emph{compositional expressivity}.
For direct human control, each discrete skill should have a predictable meaning. Pressing the same button should generally produce the same type of behavior.
At the same time, we intentionally keep the skill set small to preserve a simple interface, which limits the behaviors that individual skills can represent.

Skill transitions provide a natural way to increase expressivity without adding more buttons.
Because the low-level policy is state-conditioned, $\pi_{\mathrm{low}}(a\mid s,z)$, the behavior induced by a skill $z$ can depend on the state from which it is executed.
One skill may move the agent into a particular region of the state space, from which another skill produces a qualitatively different behavior.
This gives rise to \emph{combo behaviors} that do not appear when either skill is executed in isolation.

This does not necessarily require abandoning semantic consistency.
Instead, the state space can be viewed as containing a small number of behaviorally distinct regimes,
\begin{equation}
\mathcal{S}
=
\mathcal{S}_1
\cup
\cdots
\cup
\mathcal{S}_R,
\end{equation}
within which each skill remains relatively predictable.
Skill semantics can therefore be locally consistent within a regime while changing after the agent transitions to another regime.
This resembles a compact finite-state controller, where transitions between regimes allow the same small set of skill inputs to realize a broader behavioral repertoire.

\texttt{G1Boxing} provides a concrete example.
The backward-leaning pose forms a distinctive regime relative to ordinary locomotion and boxing states.
From regular standing states, the learned skills produce behaviors such as walking, guarding, and punching.
After entering the backward-leaning state, however, a subsequent skill can exploit the stored posture and momentum to produce a substantially stronger forward punch.

This interpretation is compatible with our mutual-information objective.
The discriminator encourages different skill codes to induce distinguishable behaviors, but does not require a skill to produce exactly the same motion from every state.
Thus, state-dependent effects can coexist with distinguishable semantics as long as the resulting behaviors remain sufficiently predictable and separable.

There is nevertheless a limit to this mechanism.
If the state space were fragmented into too many context-dependent regimes, the same button could acquire too many meanings, making the controller difficult to understand.
Our results instead suggest a useful middle ground: a small number of state-dependent regimes can increase compositional expressivity while retaining a compact and interpretable interface.
Understanding this trade-off between semantic consistency, compositionality, and human playability is an interesting direction for future work.

\section{An Alternative View of Mutual Information in Skill Discovery}
\label{app:mi_perspective}

Mutual information (MI) is often treated as a central objective for skill discovery. Here, we offer a complementary interpretation: rather than viewing MI as the primary mechanism that \emph{creates} meaningful behaviors, we view it mainly as an \emph{association mechanism} that assigns distinct behaviors to latent skill codes.

Once different skills induce behaviors that are sufficiently distinguishable for $Z$ to be inferred from the resulting behavior, the MI objective is largely satisfied. It does not by itself require those behaviors to be semantically meaningful, useful, natural, or even substantially different. Small but easily detectable behavioral differences may be sufficient. From this perspective, MI more naturally answers \emph{`which behavior should belong to which latent code?''} than \emph{`which behaviors should be discovered?''} The latter requires an additional source of behavioral structure. Below, we examine five skill-discovery approaches that use MI as a central component and ask what additional source of behavioral structure each method provides. As we will see, the nature of this additional guidance largely determines the kinds of skills that ultimately emerge.

\paragraph{DIAYN: task-agnostic exploration.}
DIAYN~\citep{eysenbach2019diversity} combines MI maximization with maximum-entropy RL. Entropy encourages broad, task-agnostic exploration, while MI partitions the resulting behavioral variation across latent codes. In simple agents, this may yield recognizable skills such as different locomotion directions. In high-dimensional agents, however, many easily distinguishable behaviors need not correspond to coherent or useful motor skills. Under this view, DIAYN relies primarily on unguided exploration to generate behavioral variation and MI to organize it.

\paragraph{ASE: imitation guidance.}
ASE~\citep{peng2022ase} combines an MI-based skill objective with adversarial imitation from motion data. The imitation objective constrains learning toward natural human motion, while MI organizes this behavioral space into a controllable latent representation. Thus, imitation largely determines learned skills, whereas MI helps make them controllable through $z$.

\paragraph{RGSD: structured latent guidance.}
Reference-Grounded Skill Discovery (RGSD)~\citep{rho2026reference} provides behavioral structure by grounding the latent representation in reference motions. Once this representation is structured, a DIAYN-style reward can encourage the policy to visit states associated with a particular reference, effectively turning an MI-form objective into an imitation-like signal. This illustrates how the behavioral effect of MI depends strongly on the representation surrounding it.

\paragraph{SDAX: locomotion-task guidance.}
SDAX~\citep{rho2025unsupervised} combines skill discovery with explicit task rewards for challenging locomotion behaviors such as leaping, climbing, and obstacle traversal. Task rewards direct exploration toward useful regions of behavior space, while the skill objective encourages different latent codes to capture distinct solutions. Skill discovery therefore acts as strategic exploration rather than diversity for its own sake.

\paragraph{GGSD: competitive-game guidance.}
\methodabbr{} follows the same broad perspective but obtains behavioral structure from competitive self-play. To defeat evolving opponents, the agent must discover effective ways to move, evade, strike, block, reposition, and interact with the environment. Because these behaviors emerge in service of winning, the resulting distinctions are naturally tied to strategically meaningful actions rather than arbitrary behavioral variation. Skills such as punching, guarding, turning, or striking a puck emerge because they are useful for winning, not because they are explicitly specified.

MI then organizes these behaviors into a compact discrete skill set. The game objective determines \emph{what kinds of behaviors are useful to discover}, while MI encourages different latent codes to represent distinguishable motor behaviors. Without the game objective, MI has no preference for semantically meaningful behaviors over arbitrary distinguishable motion; without MI, self-play need not organize effective behaviors into individually controllable skills.

This perspective suggests a common decomposition across these methods:
\begin{equation}
\underbrace{\text{behavioral guidance}}_{\text{what behaviors emerge}}
\quad+\quad
\underbrace{\text{MI-based association}}_{\text{which } z \text{ represents them}}.
\end{equation}

Under this view, the central contribution of \methodabbr{} is not a new mechanism for maximizing mutual information, but a new source of \emph{guidance for skill discovery}: competitive games provide a lightweight objective from which strategically useful and semantically meaningful motor behaviors can emerge, while MI organizes them into the discrete skill vocabulary required for direct human control.

\section{Environment Details}
\label{app:env_details}

All environments are symmetric two-player games between ``me'' and ``foe''.
Both agents receive observations with identical layouts under swapped roles,
allowing a single policy to control either side.
Following the notation in the main text, the high-level policy and critics
receive the full game state $s^{\mathrm{all}}$, while the low-level actor
receives only the controlled agent's proprioceptive observation
$s^{\mathrm{proprio}}$.
Thus, $s^{\mathrm{proprio}}$ forms a subset of $s^{\mathrm{all}}$.

Components marked with \checkmark{} are included in
$s^{\mathrm{proprio}}$ and are therefore available to the low-level actor.
The remaining components are available only through $s^{\mathrm{all}}$.
Unless noted otherwise, positions are expressed relative to the
per-environment origin and velocities in the agent's body frame.

The discriminator receives the same proprioceptive information as the
low-level actor, except that the previous-action dimensions are removed in
\texttt{AntSumo} and \texttt{G1Boxing}. 

\begin{table}[h]
\centering
\small
\caption{Observation-space dimensions.}
\label{tab:obs-summary}
\begin{tabular}{@{}lccc@{}}
\toprule
Environment
& $s^{\mathrm{proprio}}$
& $s^{\mathrm{all}}$
& \# Skills \\
\midrule
Ant Sumo       & 35 & 91  & 5 \\
Ant Fencing    & 35 & 91  & 5 \\
Franka Hockey  & 23 & 61  & 5 \\
G1 Boxing      & 86 & 194 & 6 \\
\bottomrule
\end{tabular}
\end{table}


\newpage

\subsection{Ant Sumo}
\label{app:ant-sumo}

Two quadruped ants compete in an $8\,\mathrm{m}\times8\,\mathrm{m}$ square
arena; a player loses when it is pushed out of the arena. Observation and reward are presented in Table~\ref{tab:obs-ant} and ~\ref{tab:reward_ant_sumo}, respectively. 

\begin{table}[h!]
\centering
\small
\caption{
Ant Sumo observation space.
Components marked with \checkmark{} are included in
$s^{\mathrm{proprio}}$; all components together form $s^{\mathrm{all}}$.
}
\label{tab:obs-ant}
\begin{tabular}{@{}llcc@{}}
\toprule
Group & Component & Dim. & $s^{\mathrm{proprio}}$ \\
\midrule

\multirow{8}{*}{Me}
 & Base height $z$                                       & 1 & \checkmark \\
 & Base linear velocity (body frame)                     & 3 & \checkmark \\
 & Base angular velocity (body frame)                    & 3 & \checkmark \\
 & Base yaw                                              & 1 & \checkmark \\
 & Projected gravity (body frame)                        & 3 & \checkmark \\
 & Joint positions (normalized)                          & 8 & \checkmark \\
 & Joint velocities ($\times 0.2$)                       & 8 & \checkmark \\
 & Previous action                                       & 8 & \checkmark \\
\midrule

\multirow{2}{*}{Me (global)}
 & Global position $(x,y)$                               & 2 & \\
 & Base roll                                             & 1 & \\
\midrule

Relative
 & Foe position relative to me (world frame, $xyz$)      & 3 & \\
\midrule

\multirow{9}{*}{Foe}
 & Global position $(x,y)$                               & 2 & \\
 & Base height $z$                                       & 1 & \\
 & Base linear velocity (body frame)                     & 3 & \\
 & Base angular velocity (body frame)                    & 3 & \\
 & Base yaw and roll                                     & 2 & \\
 & Projected gravity (body frame)                        & 3 & \\
 & Joint positions (normalized)                          & 8 & \\
 & Joint velocities ($\times 0.2$)                       & 8 & \\
 & Previous action                                       & 8 & \\
\midrule

\multirow{6}{*}{Strategic}
 & Planar distance to foe                                & 1 & \\
 & Foe relative position $(x,y)$ in me's body frame      & 2 & \\
 & Foe relative velocity $(x,y)$ in me's body frame      & 2 & \\
 & Normalized distances to the four arena walls          & 4 & \\
 & Relative heading $(\sin\Delta\psi,\cos\Delta\psi)$    & 2 & \\
 & Normalized episode progress $t/T$                     & 1 & \\
\midrule

\multicolumn{2}{r}{\textbf{Total $s^{\mathrm{all}}$}}
& \textbf{91} & \\
\multicolumn{2}{r}{\textbf{Total $s^{\mathrm{proprio}}$}}
& \textbf{35} & \\

\bottomrule
\end{tabular}
\end{table}

\begin{table}[h!]
\centering
\caption{Reward terms for Ant Sumo.}
\label{tab:reward_ant_sumo}
\begin{tabularx}{\linewidth}{l L l r}
\toprule
\textbf{Term} & \textbf{Description} & \textbf{Expression} & \textbf{Weight} \\
\midrule
Approach & Decrease in distance to the opponent
& $d_{t-1} - d_t$ & 2.0 \\
Push & Decrease in the opponent's distance to the nearest edge
& $w_{t-1} - w_t$ & 2.0 \\
Win & Opponent leaves the arena or falls (once, terminal)
& $\mathbf{1}[\text{win}]$ & $+20$ \\
Lose & Opponent wins
& $\mathbf{1}[\text{lose}]$  & $-20$ \\
Timeout & Draw at 10\,s (once, terminal)
& $\mathbf{1}[\text{draw}]$ & $-3$ \\
\midrule
\multicolumn{3}{l}{Alive, escape, upright, action, joint-velocity} & 0 \\
\bottomrule
\end{tabularx}
\end{table}

\newpage
\subsection{Ant Fencing}
\label{app:ant-fencing}

Ant Fencing uses the same observation space as Ant Sumo
(Table~\ref{tab:obs-ant}).
The game uses the same arena and agent configuration, but a player
additionally loses when the opponent's front foot touches its torso with
sufficient contact force. The observation space is identical to that of Ant Sumo. Reward terms are presented in Table~\ref{tab:reward_ant_fencing}.

\begin{table}[h!]
\centering
\caption{Reward terms for Ant Fencing.
The reward is identical to Ant Sumo except that the push term is disabled.
Arena $8\times8$\,m, 10\,s episodes, random initial heading.}
\label{tab:reward_ant_fencing}
\begin{tabularx}{\linewidth}{l L l r}
\toprule
\textbf{Term} & \textbf{Description} & \textbf{Expression} & \textbf{Weight} \\
\midrule
Approach & Decrease in distance to the opponent
& $d_{t-1} - d_t$ & 2.0 \\
Win & Front foot hits the opponent's torso ($\|F\| > 1$\,N),
or opponent leaves the arena / falls
& $\mathbf{1}[\text{win}]$ & $+20$ \\
Lose & Opponent's front foot hits the agent's torso,
or agent leaves the arena / falls
& $\mathbf{1}[\text{lose}]$ & $-20$ \\
Timeout & Draw at 10\,s
& $\mathbf{1}[\text{timeout}]$ & $-3$ \\
\bottomrule
\end{tabularx}
\end{table}

\subsection{Franka Air Hockey}
\label{app:franka-hockey}

Two 7-DoF Franka arms face each other across an air-hockey table and attempt
to shoot a puck into the opponent's goal; the first player to score two goals
wins the match.
All positions and velocities are expressed in the agent's own table frame,
whose $+x$ axis points toward the opponent's goal.
For the second player, the world $x$ and $y$ axes are negated so that both
players observe the game from the same viewpoint.
Joint velocities are scaled by $0.1$.
The end-effector (EE) yaw is measured relative to the agent's facing direction
and encoded as $(\sin\psi,\cos\psi)$.
The low-level actor observes only the controlled arm state, while the puck,
opponent, and match state are included only in $s^{\mathrm{all}}$.

\begin{table}[h!]
\centering
\small
\caption{
Franka Air Hockey observation space.
Components marked with \checkmark{} are included in
$s^{\mathrm{proprio}}$; all components together form $s^{\mathrm{all}}$.
}
\label{tab:obs-franka}
\begin{tabular}{@{}llcc@{}}
\toprule
Group & Component & Dim. & $s^{\mathrm{proprio}}$ \\
\midrule

\multirow{6}{*}{Me}
 & Arm joint positions                                   & 7 & \checkmark \\
 & Arm joint velocities ($\times 0.1$)                   & 7 & \checkmark \\
 & EE position (table frame)                             & 3 & \checkmark \\
 & EE linear velocity (table frame)                      & 3 & \checkmark \\
 & EE yaw $(\sin\psi,\cos\psi)$                          & 2 & \checkmark \\
 & EE yaw rate                                           & 1 & \checkmark \\
\midrule

\multirow{5}{*}{Puck}
 & Puck position (table frame)                           & 3 & \\
 & Puck linear velocity (table frame)                    & 3 & \\
 & Puck angular velocity (table frame)                   & 3 & \\
 & Puck position relative to me's EE                     & 3 & \\
 & Foe goal center relative to puck $(x,y)$              & 2 & \\
\midrule

\multirow{4}{*}{Foe}
 & Foe EE position (table frame)                         & 3 & \\
 & Foe EE linear velocity (table frame)                  & 3 & \\
 & Foe arm joint positions                               & 7 & \\
 & Foe arm joint velocities ($\times 0.1$)               & 7 & \\
\midrule

\multirow{4}{*}{Match}
 & Me score (normalized by score-to-win)                 & 1 & \\
 & Foe score (normalized by score-to-win)                & 1 & \\
 & Round progress                                        & 1 & \\
 & Normalized match progress $t/T$                       & 1 & \\
\midrule

\multicolumn{2}{r}{\textbf{Total $s^{\mathrm{all}}$}}
& \textbf{61} & \\
\multicolumn{2}{r}{\textbf{Total $s^{\mathrm{proprio}}$}}
& \textbf{23} & \\

\bottomrule
\end{tabular}
\end{table}

\begin{table}[h!]
\centering
\caption{Reward terms for Franka Air Hockey.
A match lasts 12\,s and is won by the first player to score two goals;
if the puck rests for 3\,s inside (0.5\,s outside) the workspace,
the match ends with the current score.
$x^{p}$ is the puck position along the axis toward the opponent's goal.}
\label{tab:reward_franka_hockey}
\begin{tabularx}{\linewidth}{l L l r}
\toprule
\textbf{Term} & \textbf{Description} & \textbf{Expression} & \textbf{Weight} \\
\midrule
Puck progress & Forward puck displacement toward the opponent's goal
(backward motion ignored)
& $\max(0,\, x^{p}_t - x^{p}_{t-1})$ & 10.0 \\
Goal & Agent scores (once per goal)
& $\mathbf{1}[\text{goal}]$ & $+40$ \\
Concede & Opponent scores (once per goal)
& $\mathbf{1}[\text{concede}]$ & $-40$ \\
Match win & Two goals first, or leading at timeout / idle-puck termination
& $\mathbf{1}[\text{match win}]$ & $+80$ \\
Match lose & Mirror of match win
& $\mathbf{1}[\text{match lose}]$ & $-80$ \\
Draw & Tied at termination
& $\mathbf{1}[\text{draw}]$ & 0 \\
\midrule
\multicolumn{3}{l}{Action, workspace penalties} & 0 \\
\bottomrule
\end{tabularx}
\end{table}

\subsection{G1 Boxing}
\label{app:g1-boxing}

Two Unitree G1 humanoids with 23 actuated joints compete in a
$10\,\mathrm{m}\times10\,\mathrm{m}$ arena.
Successful strikes to the opponent's torso or head reduce its hit points (HP),
and the player with more remaining HP at timeout wins.

\begin{table}[h!]
\centering
\small
\caption{
G1 Boxing observation space.
Components marked with \checkmark{} are included in
$s^{\mathrm{proprio}}$; all components together form $s^{\mathrm{all}}$.
}
\label{tab:obs-g1}
\begin{tabular}{@{}llcc@{}}
\toprule
Group & Component & Dim. & $s^{\mathrm{proprio}}$ \\
\midrule

\multirow{9}{*}{Me}
 & Base yaw                                              & 1 & \checkmark \\
 & Base height $z$                                       & 1 & \checkmark \\
 & Base linear velocity (body frame)                     & 3 & \checkmark \\
 & Base angular velocity (body frame)                    & 3 & \checkmark \\
 & Projected gravity (body frame)                        & 3 & \checkmark \\
 & Glove positions (heading frame, $2\times3$)           & 6 & \checkmark \\
 & Joint positions                                       & 23 & \checkmark \\
 & Joint velocities ($\times0.2$)                        & 23 & \checkmark \\
 & Previous action                                       & 23 & \checkmark \\
\midrule

Me (global)
 & Global position $(x,y)$                               & 2 & \\
\midrule

Relative
 & Foe position relative to me (heading frame, $xyz$)    & 3 & \\
\midrule

\multirow{2}{*}{Foe gloves}
 & Foe glove positions (me's torso frame, $2\times3$)    & 6 & \\
 & Foe glove velocities (me's torso frame, $2\times3$)   & 6 & \\
\midrule

\multirow{10}{*}{Foe}
 & Global position $(x,y)$                               & 2 & \\
 & Base yaw                                              & 1 & \\
 & Base height $z$                                       & 1 & \\
 & Base linear velocity (body frame)                     & 3 & \\
 & Base angular velocity (body frame)                    & 3 & \\
 & Projected gravity (body frame)                        & 3 & \\
 & Glove positions (heading frame, $2\times3$)           & 6 & \\
 & Joint positions                                       & 23 & \\
 & Joint velocities ($\times0.2$)                        & 23 & \\
 & Previous action                                       & 23 & \\
\midrule

\multirow{3}{*}{Match}
 & Me HP (normalized by initial HP)                      & 1 & \\
 & Foe HP (normalized by initial HP)                     & 1 & \\
 & Normalized time remaining $1-t/T$                    & 1 & \\
\midrule

\multicolumn{2}{r}{\textbf{Total $s^{\mathrm{all}}$}}
& \textbf{194} & \\
\multicolumn{2}{r}{\textbf{Total $s^{\mathrm{proprio}}$}}
& \textbf{86} & \\

\bottomrule
\end{tabular}
\end{table}

\begin{table}[h!]
\centering
\caption{Reward terms for G1 Boxing, adapted from Won et al.~\cite{won2021control}.
Arena $10\times10$\,m, 15\,s episodes, initial HP $H_0 = 10^{5}$ per player.
Punch damage is $150\,\bar v_\perp^{2}$, where $\bar v_\perp$ is the glove's
closing velocity normal to the struck face clipped at 12\,m/s
(head hits $\times1.5$); it drains the victim's HP.
The outcome is decided by HP alone; a fall or arena exit voids the round.
$\hat{\mathbf{u}}$ is the unit vector toward the opponent,
$\mathbf{f}_{\mathrm{torso}}, \mathbf{f}_{\mathrm{pelvis}}$ are the forward
axes of the torso and pelvis,
$\mathbf{v}_\perp$ is the root XY velocity rotated by $90^\circ$
(speed clipped at 1\,m/s),
$\tilde{q}$ is the joint position normalised to $[-1,1]$ within its soft
limits, and $F_{ij}$ are self-contact forces.}
\label{tab:reward_g1_boxing}
\begin{tabularx}{\linewidth}{l L l r}
\toprule
\textbf{Term} & \textbf{Description} & \textbf{Expression} & \textbf{Weight} \\
\midrule
Punch & HP drained from the opponent minus HP lost
& $(\Delta H^{\mathrm{opp}}_t - \Delta H^{\mathrm{me}}_t)/H_0$ & 125 \\
Facing & Torso and pelvis facing the opponent, in $[-2,2]$
& $\mathbf{f}_{\mathrm{torso}}\!\cdot\!\hat{\mathbf{u}}
 + \mathbf{f}_{\mathrm{pelvis}}\!\cdot\!\hat{\mathbf{u}}$ & 0.03 \\
Facing velocity & Penalises moving sideways relative to the body's facing
& $-\big(|\mathbf{f}_{\mathrm{torso}}\!\cdot\!\mathbf{v}_\perp|
 + |\mathbf{f}_{\mathrm{pelvis}}\!\cdot\!\mathbf{v}_\perp|\big)$ & 0.15 \\
Approach & Decrease in distance to the opponent; zero when $d_t \le 1$\,m
& $d_{t-1} - d_t$ & 3.0 \\
\midrule
Joint velocity & Squared joint velocity, clipped at $\pm 30$\,rad/s
& $-\sum_j \mathrm{clip}(\dot q_j)^2$ & $3\times10^{-5}$ \\
Joint limit & Excess beyond 80\,\% of the soft joint range, clipped at 0.3
& $-\sum_j \mathrm{clip}\big(\max(|\tilde q_j|-0.8,\,0)\big)^2$ & 0.5 \\
Action rate & Squared change of the action
& $-\|\mathbf{a}_t-\mathbf{a}_{t-1}\|^2$ & 0.005 \\
Self-contact & Sum of self-collision forces, clipped at 50\,N per pair
& $-\sum_{(i,j)} \mathrm{clip}(\|F_{ij}\|)$ & $3\times10^{-4}$ \\
\midrule
Win / Lose & Opponent's HP reaches zero first, or higher HP at timeout
(mirror for lose)
& $\mathbf{1}[\text{win}] \,/\, \mathbf{1}[\text{lose}]$
& $+30 \,/\, -30$ \\
Draw & Simultaneous HP-out or equal HP at timeout
& $\mathbf{1}[\text{draw}]$ & $-5$ \\
Arena out & Agent falls (torso below 0.65\,m) or leaves the arena
& $\mathbf{1}[\text{arena out}]$ & $-3$ \\

\bottomrule
\end{tabularx}
\end{table}

\clearpage

\section{Hyperaparameters}
\label{app:hyperparam}

We provide the hyperparameters used in our experiments.

\begin{table}[h!]
\centering
\caption{Training hyper-parameters of the hierarchical self-play policies.}
\label{tab:hyperparams}
\small
\begin{tabular}{l c c c c}
\toprule
\textbf{Hyper-parameter} & \textbf{Ant Sumo} & \textbf{Ant Fencing} & \textbf{Franka Hockey} & \textbf{G1 Boxing} \\
\midrule
\multicolumn{5}{l}{\textit{PPO}} \\
Parallel environments & \multicolumn{4}{c}{4096} \\
Rollout length (steps / env) & 32 & 32 & 24 & 32 \\
Training iterations & \multicolumn{4}{c}{70\,000} \\
Learning epochs / mini-batches & \multicolumn{4}{c}{5 / 4} \\
Learning rate & \multicolumn{4}{c}{$1\times10^{-4}$} \\
Discount $\gamma$ & 0.99 & 0.99 & 0.995 & 0.99 \\
GAE $\lambda$ & \multicolumn{4}{c}{0.95} \\
Clip ratio $\epsilon$ & \multicolumn{4}{c}{0.2} \\
Value loss coefficient (clipped) & \multicolumn{4}{c}{1.0} \\
Max gradient norm & \multicolumn{4}{c}{1.0} \\
Entropy coefficient, low-level & $5\times10^{-4}$ & $5\times10^{-4}$ & $1\times10^{-3}$ & $5\times10^{-4}$ \\
Entropy coefficient, high-level & 0.02 & 0.02 & 0.02 & 0.05 \\
Initial action std, low-level & 1.0 & 1.0 & 1.0 & 0.2 \\
Action clipping & 10 & 10 & 10 & 1 \\
Control frequency (Hz) & 60 & 60 & 60 & 30 \\
Seed & \multicolumn{4}{c}{42} \\
\midrule
\multicolumn{5}{l}{\textit{Hierarchy and skill discovery}} \\
Number of skills $K$ & 5 & 5 & 5 & 6 \\
Skill duration & \multicolumn{4}{c}{10} \\
MI reward scale $\beta$ & 0.02 & 0.02 & 0.005 & 0.05 \\
Discriminator learning rate & \multicolumn{4}{c}{$1\times10^{-4}$} \\
\midrule
\multicolumn{5}{l}{\textit{Self-play opponent pool}} \\
Warm-up iterations (mirror self-play) & 0 & 0 & 2000 & 2000 \\
Opponent snapshot interval (iterations) & 2000 & 2000 & 2000 & 1000 \\
Opponent resampling interval (iterations) & \multicolumn{4}{c}{200} \\
Opponent groups & 10 & 10 & 4 & 10 \\
Fraction of groups using the latest policy & 0.6 & 0.6 & 0.6 & 0.3 \\
\bottomrule
\end{tabular}
\end{table}

\paragraph{Network architecture.}
All networks are multilayer perceptrons with ELU activations. For the Ant Sumo, Ant Fencing, and G1 Boxing tasks, the high-level policy has three hidden layers of [512, 512, 256] units and outputs a categorical distribution over the $K$ skills; the low-level policy and both critics (one for each level) have four hidden layers of [512, 512, 256, 128] units, and the skill discriminator has two hidden layers of 256 units. For Franka Hockey, the high-level policy, the low-level policy, the critics, and the discriminator all use four hidden layers of [512, 512, 512, 256] units. The low-level policy receives only the agent's own proprioceptive state together with the one-hot skill and outputs the mean of a Gaussian over joint actions with a learned state-independent standard deviation; for G1 Boxing the mean is additionally bounded with a $\tanh$. 

\section{VLM-Based Skill Diversity Metric}
\label{app:vlm_diversity}

\paragraph{Pipeline.}
For every trained policy we record one clip per skill (Ant: 5, Franka: 5, G1: 6 skills; for METRA a skill is the one-hot unit vector along each latent axis). Each clip is a single-environment rollout that starts from a reset state and runs for 5\,s, rendered at $1920\times1080$. The clip is captioned by a vision-language model (VLM), the caption is embedded with a text-embedding model, and the diversity score of a policy is the mean pairwise cosine distance
\[
D=\frac{1}{\binom{K}{2}}\sum_{i<j}\bigl(1-\cos(\mathbf{e}_i,\mathbf{e}_j)\bigr)
\]
over the $K$ skill embeddings $\mathbf{e}_i$ (L2-normalized). The prompt asks the VLM to answer \emph{no semantic meaning} when a clip shows no interpretable behavior; every pair containing such a clip is assigned distance $0$, so uninterpretable skills do not add diversity. Scores are averaged over 3 training seeds (mean $\pm$ standard deviation).

\paragraph{Models and settings.}
Captioning uses \texttt{gemini-3.6-flash} (temperature $0.1$, at most 1024 output tokens). The video is sampled by the API at 5\,fps over a fixed segment of the clip: Ant 1--4\,s, Franka 2--4\,s, G1 0--5\,s. The segment skips the initial settling phase after the reset where the robot has not yet started to move. Captions are embedded with \texttt{gemini-embedding-001} (task type \texttt{SEMANTIC\_SIMILARITY}, 3072 dimensions). The marker \emph{no semantic meaning} is matched case-insensitively as a substring, since the model occasionally wraps it in quotes or a full sentence.

\paragraph{Prompts.}
One prompt is used per robot; the same prompt is applied to every method on that robot.

\begin{itemize}
  \item \textbf{Ant.}
  \textit{``This is a rendered clip of a simulated robot. Describe the semantic meaning of the robot's behavior in one sentence. If you cannot think of a meaningful downstream task for which this behavior could be useful, say `no semantic meaning'.''}

  \item \textbf{Franka.}
  \textit{``This is a rendered clip of a simulated robot arm. Describe which region of the table surface the robot arm moves over. If the robot arm moves too far away from the table, or would otherwise be unlikely to meaningfully interact with objects on the table, say `no semantic meaning'.''}

  \item \textbf{G1.}
  \textit{``This is a rendered clip of a humanoid robot with a sphere attached to each hand. Describe the semantic meaning of the robot's behavior in one sentence. If the behavior has no clear semantic meaning, respond with exactly `no semantic meaning'.''}
\end{itemize}

\paragraph{Example captions.}
Ant (ours): \textit{``The quadruped robot performs a self-righting maneuver to flip itself over and stand back up on its feet.''}, \textit{``The robot is rotating in place.''}, \textit{``The robot is crawling forward.''}
Franka (ours): \textit{``The robot arm remains stationary over the bottom-right region of the table surface.''}
G1 (ours): \textit{``The humanoid robot is walking backwards.''}

\end{document}

%% file: math_commands.tex
\usepackage{amsmath,amsfonts,bm}

\newcommand{\methodabbr}{GGSD}

\def\eqref#1{equation~\ref{#1}}

\def\1{\bm{1}}

\DeclareMathAlphabet{\mathsfit}{\encodingdefault}{\sfdefault}{m}{sl}
\SetMathAlphabet{\mathsfit}{bold}{\encodingdefault}{\sfdefault}{bx}{n}

\DeclareMathOperator*{\argmin}{arg\,min}